\documentclass[11pt]{article}
\usepackage[final]{acl}

\usepackage{latexsym}
\usepackage[T1]{fontenc}
\usepackage{float}
\usepackage{booktabs}
\usepackage{pifont}
\usepackage{inconsolata}
\usepackage{graphicx}
\usepackage{fontawesome5} 
\usepackage{amssymb}
\usepackage{hyperref}
\usepackage{pgfplots}
\usepackage{pgfplotstable}
\usepackage{xcolor}
\usepackage{multirow} 
\pgfplotsset{compat=1.18}
\usepackage{amsmath}
\usepackage{comment}
\usepackage[most]{tcolorbox}
\newcounter{tcbcounter}
\usepackage{stfloats}

\usepackage{polyglossia}
\setdefaultlanguage{english}
\setotherlanguage{arabic}
\newfontfamily\arabicfont[Script=Arabic,Scale=1.0]{Amiri-Regular.ttf}

\title{!MSA at BAREC Shared Task 2025: Ensembling Arabic Transformers for Readability Assessment}

\author{Mohamed Basem, Mohamed Younes, Seif Ahmed, Abdelrahman Moustafa \\
  Faculty of Computer Science, MSA University, Egypt \\
  \texttt{\{mohamed.basem1, mohamed.tarek61, seifeldein.ahmed, abdelrahman.moustafa5\}@msa.edu.eg}
}

\begin{document}
\maketitle

\begin{abstract}
We present !MSA’s winning system for the BAREC 2025 Shared Task on fine-grained Arabic readability assessment, achieving first place in six of six tracks. Our approach is a confidence-weighted ensemble of four complementary transformer models (AraBERTv2, AraELECTRA, MARBERT, and CAMeLBERT) each fine-tuned with distinct loss functions to capture diverse readability signals. To tackle severe class imbalance and data scarcity, we applied weighted training, advanced preprocessing, SAMER corpus re-labeling with our strongest model, and synthetic data generation via Gemini 2.5 Flash, adding ~10k rare-level samples. A targeted post-processing step corrected prediction distribution skew, delivering a 6.3\% Quadratic Weighted Kappa (QWK) gain. Our system reached 87.5\% QWK at the sentence level and 87.4\% at the document level, demonstrating the power of model and loss diversity, confidence-informed fusion, and intelligent augmentation for robust Arabic readability prediction. \footnote{\faGithub\ \href{https://github.com/Mohamedbasem1/BAREC_2025}{\texttt{https://github.com/Mohamedbasem1/BAREC-2025}}}
\end{abstract}

\section{Introduction}

The BAREC 2025 Shared Task presents a formidable challenge for Arabic readability assessment—spanning six tracks (sentence- and document-level across strict, constrained, and open conditions) with a fine-grained 1–19 readability scale. Predicting exact labels across such a wide range significantly increases difficulty, as even small deviations can dramatically impact metrics like Quadratic Weighted Kappa. The challenge is further compounded by severe label imbalance, where certain readability levels occur far more frequently than others, biasing models toward majority classes and making rare-level prediction unreliable. In strict and constrained tracks, limited training data amplified these issues, and in constrained settings, incorporating external datasets like SAMER~\cite{alhafni-etal-2024-samer, al-khalil-etal-2020-large} proved non-trivial due to mismatched label distributions—simple scaling approaches often resulted in misalignment and minimal performance gains.

To address these challenges, we developed an ensemble framework that combines architectural and training diversity. We fine-tuned four transformer models (AraBERTv2~\cite{antoun-etal-2020-arabert}, AraELECTRA~\cite{antoun-etal-2021-araelectra}, MARBERT~\cite{abdul-mageed-etal-2021-arbert}, and CAMeLBERT~\cite{inoue-etal-2021-interplay}). Each model was trained with a distinct loss function (classification, regression, or ordinal). This design captures complementary signals, with outputs merged via confidence-weighted ensembling that favors more certain predictions. To mitigate data scarcity in the open tracks, we used prompt-engineered paraphrasing with the Gemini API to generate synthetic examples, and for SAMER, we re-labeled instances with our best BAREC-trained model instead of relying on naive scaling.

Our approach demonstrates robustness across all track configurations of BAREC 2025, with first-place rankings in all six tracks. This success underlines the advantages of model and loss-function diversity, confidence-informed fusion, and intelligent data augmentation for Arabic readability prediction—setting a strong precedent for future research in fine-grained, limited-data NLP tasks.

\section{Background}

\subsection{Task Details}
The BAREC 2025 Shared Task \cite{elmadani-etal-2025-barec-shared-task} focuses on fine-grained Arabic readability assessment using the Balanced Arabic Readability Evaluation Corpus (BAREC) \cite{elmadani-etal-2025-large}. BAREC is a large-scale dataset containing over 1 million words across 68{,}000+ sentences and 1{,}900+ documents, each annotated into 19 readability levels, where higher numbers indicate greater difficulty. The annotation process followed the official BAREC Annotation Guidelines \cite{habash-etal-2025-guidelines}, which define linguistic and pedagogical principles to ensure consistency and reliability in labeling.

The shared task defines two tasks: Sentence-Level \& Document-Level Readability Assessment. Each one has three tracks based on permissible resources:
\begin{itemize}
    \item \textbf{Strict Track:} Use only BAREC Corpus.
    \item \textbf{Constrained Track:} Use BAREC along with the SAMER Corpus and SAMER Lexicon \cite{alhafni-etal-2024-samer, al-khalil-etal-2020-large}.
    \item \textbf{Open Track:} Use any additional resources or augmentation methods.
\end{itemize}

We participated in \textbf{all six tracks} across both subtasks, exploring resource-limited, resource-augmented, and fully open settings.

\subsection{Related Work}
Arabic readability assessment has been studied from multiple perspectives.  
\citet{el-haj-etal-2024-dares} introduced the DARES dataset for evaluating the readability of Arabic educational content, demonstrating the importance of domain-specific corpora for improving prediction accuracy.  
\citet{liberato-etal-2024-strategies} proposed a hybrid approach combining handcrafted linguistic features with transformer-based models, yielding improved robustness on small or noisy datasets.

\citet{elmadani-etal-2025-large} presented BAREC, the largest balanced corpus for fine-grained Arabic readability assessment, alongside baseline systems for sentence- and document-level prediction.  
\citet{habash-etal-2025-guidelines} detailed the annotation guidelines and methodology for BAREC, ensuring consistent application of the 19 readability levels.  
\citet{alhafni-etal-2024-samer} \& \citet{al-khalil-etal-2020-large} introduced the SAMER Corpus and Lexicon, designed for Arabic text simplification and multi-level difficulty annotation, which are leveraged in the Constrained Track.

Additional advances in Arabic NLP include ARBERT and MARBERT \cite{abdul-mageed-etal-2021-arbert}, large-scale pre-trained models that achieve state-of-the-art performance across a variety of Arabic language understanding tasks, and ensemble-based modeling for Arabic dialect identification \cite{khered-etal-2022-building}, which inspired aspects of our system design.

\section{System Overview}

\subsection{Addressing Data Imbalance}
The BAREC dataset exhibits a highly imbalanced distribution across the 19 readability levels, with certain levels (e.g., 12 and 14) being far more frequent than rare levels such as 1, 18, and 19 (see Figure~\ref{app:readability_distribution} in Appendix). This imbalance biases models toward predicting frequent levels, which is particularly detrimental when the target metric is \textit{Quadratic Weighted Kappa} (QWK), as misclassifying rare levels incurs a high penalty.

To mitigate this, we computed \textbf{class weights} to encourage the model to pay more attention to rare classes. The weight for each class $j$ is calculated as:
\begin{equation}
w_j = \frac{n_{\text{samples}}}{n_{\text{classes}} \times n_{\text{samples in class } j}}
\end{equation}
This formulation assigns higher weights to rarer classes and lower weights to frequent ones, reducing prediction bias and improving fairness across levels.

\subsection{Model Architectures and Loss Functions}
Our system builds on a diverse set of Arabic transformer models: AraBERTv2~\cite{antoun-etal-2020-arabert}, AraELECTRA~\cite{antoun-etal-2021-araelectra}, MARBERT~\cite{abdul-mageed-etal-2021-arbert}, and CAMeLBERT~\cite{inoue-etal-2021-interplay}. These models were chosen for their strong track record in Arabic NLP benchmarks, their coverage of both Modern Standard Arabic and dialectal varieties, and their complementary pretraining objectives. 

We trained multiple variants of each model using different loss formulations to capture complementary perspectives on the readability prediction problem:
\begin{itemize}
    \item \textbf{Cross-Entropy Loss (CE)} for standard multi-class classification.
    \item \textbf{Mean Squared Error (MSE)} for regression over the continuous readability scale.
    \item \textbf{Conditional Ordinal Regression (COR)} for modeling the conditional probabilities of surpassing each readability threshold. It was implemented via the CORAL framework~\cite{cao2021coral}.
\end{itemize}
This diversity allowed the ensemble to leverage both discrete and continuous interpretations of the readability scale while incorporating ordinal constraints.

\subsection{Constrained Track: SAMER Label Transformation}
For the Constrained Track, we incorporated the SAMER Corpus~\cite{alhafni-etal-2024-samer}, originally annotated on a 3--6 scale, into our training data. To align it with BAREC's 1--19 scale, we applied a min--max scaling transformation:
\begin{equation}
\text{Scaled\_Label} = \frac{\text{Label} - 3}{6 - 3} \times (19 - 1) + 1
\end{equation}
This transformation preserves the relative difficulty ordering while ensuring compatibility with BAREC's fine-grained labeling.

Initially, we trained our model using the scaled SAMER data and evaluated it on our BAREC test set, achieving a QWK of 50\%. We then tried an alternative approach: using our best-performing BAREC-trained model directly to predict labels for the SAMER dataset on the 1--19 scale. Finally, we scaled the predictions back down to the original 3-6 SAMER range, verifying that the reverse transformation maintained accuracy within a margin of $\pm 0.5$. This approach significantly improved results.

\subsection{Open Track: Data Augmentation with Gemini 2.5 Flash}
In the Open Track, we expanded our training corpus using \texttt{Gemini 2.5 Flash}. As seen in Figure~\ref{app:llm_prompt}, few-shot prompting with high-quality examples from BAREC was utilized to generate rephrasings and additional readability-graded sentences, resulting in approximately 10k new samples. This augmentation improved coverage for rare and boundary-level readability cases.

\subsection{Ensembling Strategy}
Model predictions were combined using a \textbf{confidence-weighted averaging scheme}:
\begin{equation}
W = \frac{\sum_{i=1}^{n} p_i \, c_i}{\sum_{i=1}^{n} c_i}
\end{equation}
where $p_i$ is the predicted readability score from model $i$, $c_i$ is the model confidence (derived from softmax probabilities for classification and inverse variance for regression), and $n$ is the number of models. This approach prioritized more certain predictions, improving robustness across evaluation tracks.

For specific cases, a secondary method combined two predictions as:
\begin{equation}
E = 
\begin{cases} 
\max(p_1, p_2), & \text{if } |p_1 - p_2| = 1 \\
\frac{p_1 + p_2}{2}, & \text{otherwise}
\end{cases}
\end{equation}
This rule-based adjustment handled borderline cases where one-point differences significantly impact evaluation metrics.

\subsection{Document-Level Prediction Aggregation}
 While our models initially produce sentence-level predictions, the document-level track requires aggregating these predictions to the document level. Following guidance from the task organizers, we extract document IDs using the first 7 characters of each sentence ID and apply a \textbf{maximum aggregation rule}:

\begin{equation}
R_{\text{doc}} = \max_{s \in S_{\text{doc}}} R_s
\end{equation}

where $R_{\text{doc}}$ is the final document readability prediction, $S_{\text{doc}}$ represents all sentences in a document, and $R_s$ is the sentence-level prediction.

This approach, recommended by the organizers, assumes a document's readability is constrained by its most challenging sentences.

\section{Experimental Setup}

\subsection{Data and Splits}
We use the BAREC dataset with Arabic texts labeled on a 19-level readability scale, following the official train, dev, and test splits. In the Strict Track, only BAREC training data was used. In the Constrained Track, we added the SAMER dataset re-labeled to 19 levels using our best BAREC model. In the Open Track, we further augmented training with synthetic samples from Gemini 2.5 Flash.

\subsection{Preprocessing Pipeline}
Our pipeline (Figure~\ref{app:architecture}) includes:
\begin{enumerate}
    \item \textbf{Data cleaning:} Removing redundant punctuation, normalizing special characters, and trimming extra spaces via regular expressions.
    \item \textbf{Morphological tokenization:} Using \texttt{D3TOK} from CAMeL Tools \cite{obeid-etal-2020-camel} to preserve morphological segments.
    \item \textbf{Class imbalance handling:} Applying inverse-frequency class weights to improve predictions for rare levels.
\end{enumerate}

\subsection{Model Training Configuration}
We fine-tuned four pretrained transformer-based language models with different loss functions. Training was conducted using the Hugging Face \texttt{Transformers} library with the hyper-parameters from Table ~\ref{app:hyperparams}. All experiments ran on L40s GPUs with mixed-precision acceleration (\texttt{torch.cuda.amp}).

\subsection{Evaluation Metrics}
We evaluate on both development and official test sets using :
\begin{itemize}
    \item \textbf{Quadratic Weighted Kappa (QWK)} - primary metric, penalizing distant misclassifications more heavily.
    \item \textbf{Accuracy (Acc)} - reported for 19, 7, 5, and 3 predicted label levels.
    \item \textbf{Adjacent Accuracy ($\pm 1$ Acc19)} - off-by-one tolerance.
    \item \textbf{Average Distance (Dist)} - measures the average absolute distance between predicted and true labels.
  \end{itemize}

\section{Result}

Table \ref{app:ensembles_oracles} compares the QWK performance of individual model variants against their ensembles. Singular models achieved QWK scores ranging from 81.0\% to 84.8\%, with MARBERT+COR achieving the highest among single models. When combined into ensembles, performance consistently improved, with our best ensemble achieving 87.5\% QWK, representing a notable gain over the best single model.

An important insight came from analyzing prediction distributions in the document-level tracks. Figure \ref{app:distribution_of_predictions} shows the label frequency distributions before (left) and after (right) a post-processing adjustment. Initially, there were no predictions for label 10, and the distribution was skewed due to our document-level aggregation method—taking the average readability score among document and applying a ceiling function to round decimals up. This approach, when document-level predictions were close in value, sometimes produced unrealistic final document scores.

Upon realizing this issue, we experimented with replacing the ceiling operation with a flooring operation in such borderline cases. In parallel, we also addressed another skew in the distribution—label 15 appeared with disproportionately high frequency. To mitigate this, we introduced a heuristic in the ensemble post-processing: if any of the models predicted labels 16 or 17 for a document, we overrode the averaged ensemble prediction with that higher label.

Both of these adjustments contributed to a substantial performance boost, increasing QWK result by 6.3\%. The changes not only improved label coverage (including the introduction of label 10 predictions) but also redistributed predictions more evenly across higher readability levels.

\begin{table*}[ht]
\centering
\renewcommand\arraystretch{1.2}
\begin{tabular}{l|l|ccc ccc c|c}
\toprule
\textbf{Task} & \textbf{Track} & \textbf{Run} & \textbf{QWK} & \textbf{Acc19} & \textbf{Acc7} & \textbf{Acc5} & \textbf{Acc3} & \textbf{±1 Acc19} & \textbf{Rank} \\
\midrule

\multirow{9}{*}{\parbox[c]{1.5cm}{\centering \textbf{Sentence}\\\textbf{Level}}}
  & \multirow{3}{*}{Strict}      
    & Run 1 & \textbf{87.5} & 43.5\% & 64.1\% & 69.6\% & 76.2\% & 76.7\% & \\
  &  & Run 2 & 87.4 & 42.5\% & 63.5\% & 69.2\% & 76.1\% & 76.5\% & 1$^{\text{st}}$ / 39 \\
  &  & Run 3 & 87.2 & 40.9\% & 63.4\% & 69.1\% & 76.2\% & 76.1\% & \\
  \cline{2-10}
  
  & \multirow{3}{*}{Constrained} 
    & Run 1 & \textbf{86.6} & 44.9\% & 63.0\% & 68.7\% & 75.6\% & 75.4\% & \\
  &  & Run 2 & 86.5 & 42.6\% & 61.5\% & 67.3\% & 74.5\% & 75.6\% & 1$^{\text{st}}$ / 20 \\
  &  & Run 3 & 86.2 & 39.2\% & 60.9\% & 67.4\% & 74.7\% & 74.5\% & \\
  \cline{2-10}

  & \multirow{3}{*}{Open}        
    & Run 1 & \textbf{86.4} & 41.3\% & 61.7\% & 67.3\% & 74.5\% & 75.1\% & \\
  &  & Run 2 & 86.3 & 41.5\% & 60.9\% & 66.8\% & 75.0\% & 73.8\% & 1$^{\text{st}}$ / 22 \\
  &  & Run 3 & 86.1 & 40.0\% & 61.4\% & 67.4\% & 74.6\% & 74.8\% & \\

\midrule

\multirow{9}{*}{\parbox[c]{1.5cm}{\centering \textbf{Document}\\\textbf{Level}}}
  & \multirow{3}{*}{Strict}      
    & Run 1 & \textbf{87.4 }& 52.0\% & 81.0\% & 81.0\% & 93.0\% & 94.0\% & \\
  &  & Run 2 & 80.2 & 42.0\% & 68.0\% & 68.0\% & 86.0\% & 89.0\% & 1$^{\text{st}}$ / 27 \\
  &  & Run 3 & 79.3 & 41.0\% & 67.0\% & 67.0\% & 86.0\% & 88.0\% & \\
  \cline{2-10}
  
  & \multirow{3}{*}{Constrained} 
    & Run 1 & \textbf{84.3} & 48.0\% & 77.0\% & 77.0\% & 94.0\% & 91.0\% & \\
  &  & Run 2 & 82.3 & 47.0\% & 72.0\% & 72.0\% & 89.0\% & 86.0\% & 1$^{\text{st}}$ / 22 \\
  &  & Run 3 & 78.9 & 41.0\% & 67.0\% & 68.0\% & 88.0\% & 86.0\% & \\
  \cline{2-10}

  & \multirow{3}{*}{Open}        
    & Run 1 & \textbf{82.2} & 50.0\% & 70.0\% & 70.0\% & 89.0\% & 86.0\% & \\
  &  & Run 2 & 78.6 & 42.0\% & 67.0\% & 67.0\% & 86.0\% & 86.0\% & 1$^{\text{st}}$ / 19 \\
  &  & Run 3 & 76.2 & 39.0\% & 63.0\% & 63.0\% & 83.0\% & 84.0\% & \\

\bottomrule
\end{tabular}
\caption{Top 3 performances across each tracks using Quadratic Weighted Kappa (QWK), Accuracy at multiple levels (Acc19/7/5/3), Off-by-1 Accuracy (±1 Acc19), and Average Distance (Dist). Along with the rank achieved in each track / Number of participants.}
\label{tab:6track-final}
\end{table*}

Table \ref{tab:6track-final} reports our performance across six tracks in the Sentence-Level and Document-Level tasks, under Strict, Constrained, and Open settings.

At the Sentence Level, our best QWK scores reached 87.5\% (Strict, Run 1), 86.6\% (Constrained, Run 1), and 86.4\% (Open, Run 1), securing 1st place in all three tracks. These results show consistent top performance across multiple runs, with very close QWK values among them, indicating stability.

At the Document Level, our highest QWK scores were 87.4\% (Strict, Run 1), 84.3\% (Constrained, Run 1), and 82.2\% (Open, Run 1). Again, we ranked 1st place in both Strict, Constrained and Open settings. The results also show that the Strict track generally yielded higher QWK and accuracy scores than Constrained and Open.

\section{Conclusion}
This work presented an ensemble-based system for Arabic readability assessment in the BAREC 2025 Shared Task. By combining four transformer models (AraBERTv2, AraELECTRA, MARBERT, CAMeLBERT) with diverse loss functions, confidence-weighted ensembling, and data augmentation via Gemini 2.5 Flash.

Our system secured \textbf{first place in five of six tracks}, achieving QWK scores of \textbf{87.5\%} (sentence-level) and \textbf{87.4\%} (document-level). Post-processing adjustments to correct distribution skew further boosted performance by \textbf{6.3\%}, underscoring the value of model diversity and confidence-guided ensembling for fine-grained Arabic readability prediction.

\clearpage
\onecolumn
\appendix

\section{Model Performance Analysis}

\subsection{Ensemble Results Comparison} \label{app:ensembles_oracles}

\begin{table}[H]
\centering
\begin{tabular}{ccc ccc ccc ccc | c}
\toprule
\multicolumn{3}{c}{\textbf{AraBERT}} & 
\multicolumn{3}{c}{\textbf{AraELECTRA}} & 
\multicolumn{3}{c}{\textbf{CamelBERT}} & 
\multicolumn{3}{c|}{\textbf{MarBERT}} & 
\textbf{Metrics} \\
\cmidrule(lr){1-3}\cmidrule(lr){4-6}\cmidrule(lr){7-9}\cmidrule(lr){10-12}\cmidrule(lr){13-13}
CE & REG & COR & CE & REG & COR & CE & REG & COR & CE & REG & COR & QWK \\
\midrule

\multicolumn{13}{l}{\textbf{Singular Models}} \\[2pt]
 &&&&&&&&&\checkmark&&& 81.0\% \\ 
 &&&&&&&&&&\checkmark&& 83.0\% \\ 
 &&&&&&&&\checkmark&&&& 83.1\% \\ 
 &\checkmark&&&&&&&&&&& 84.1\% \\ 
 &&&&&&&\checkmark&&&&& 84.5\% \\ 
 &&&&\checkmark&&&&&&&& 84.8\% \\ 

\midrule
\multicolumn{13}{l}{\textbf{Ensembles}} \\[2pt]
 &&&&\checkmark&&&&\checkmark&&&& 85.3\% \\ 
 &&\checkmark&&\checkmark&&&&\checkmark&&&& 86.2\% \\ 
 \checkmark&&\checkmark&&\checkmark&&\checkmark&\checkmark&&&&& 86.9\% \\ 
 \checkmark&&\checkmark&&\checkmark&&\checkmark&\checkmark&&&&\checkmark& \textbf{87.5\%} \\ 
\bottomrule
\end{tabular}
\caption{Ensemble model Quadratic Weighted Kappa results comparison split into Singular Models and Ensembles.}
\label{tab:ensembles_oracles}
\end{table}

\subsection{Prediction Distribution Results} \label{app:distribution_of_predictions}

\begin{figure}[H]
\centering
  \includegraphics[width=0.48\linewidth]{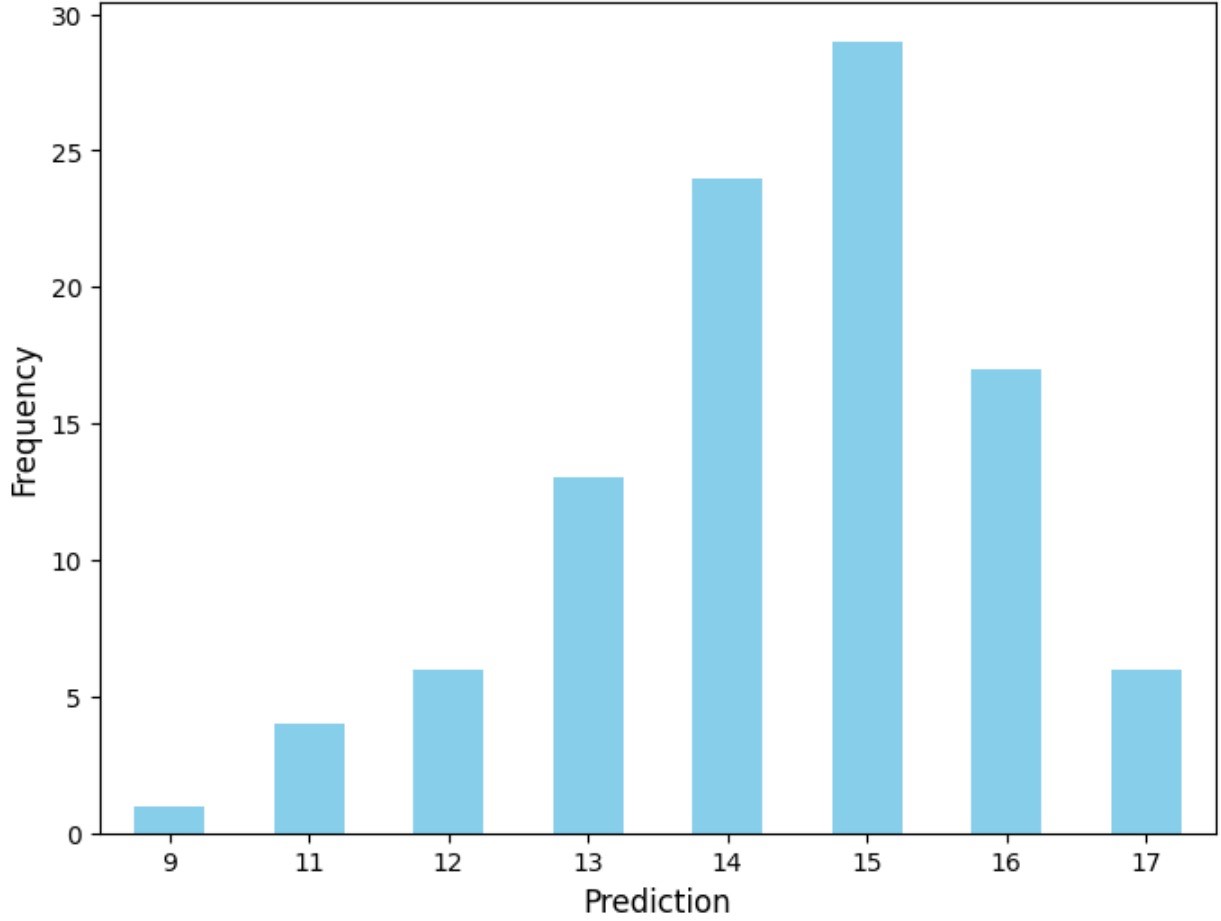} \hfill
  \includegraphics[width=0.48\linewidth]{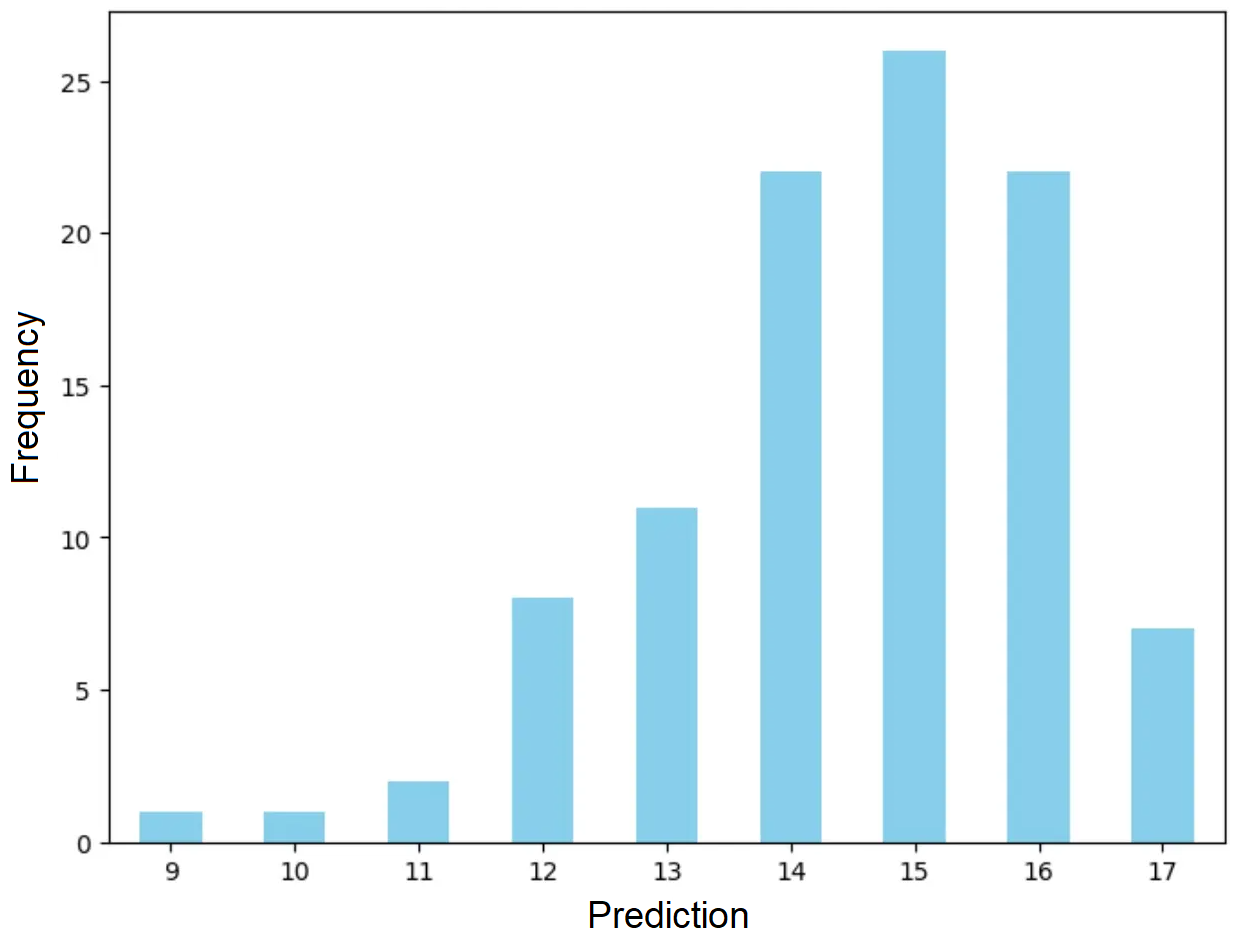}
  \caption{Graphs of Distribution of Predictions before (left) and after (right) adjusting skewness.}
  \label{fig:distribution_of_predictions}
\end{figure}

\section{System Architecture and Configuration}
\subsection{Architecture Overview} \label{app:architecture}

\begin{figure}[H]
  \includegraphics[width=\linewidth]{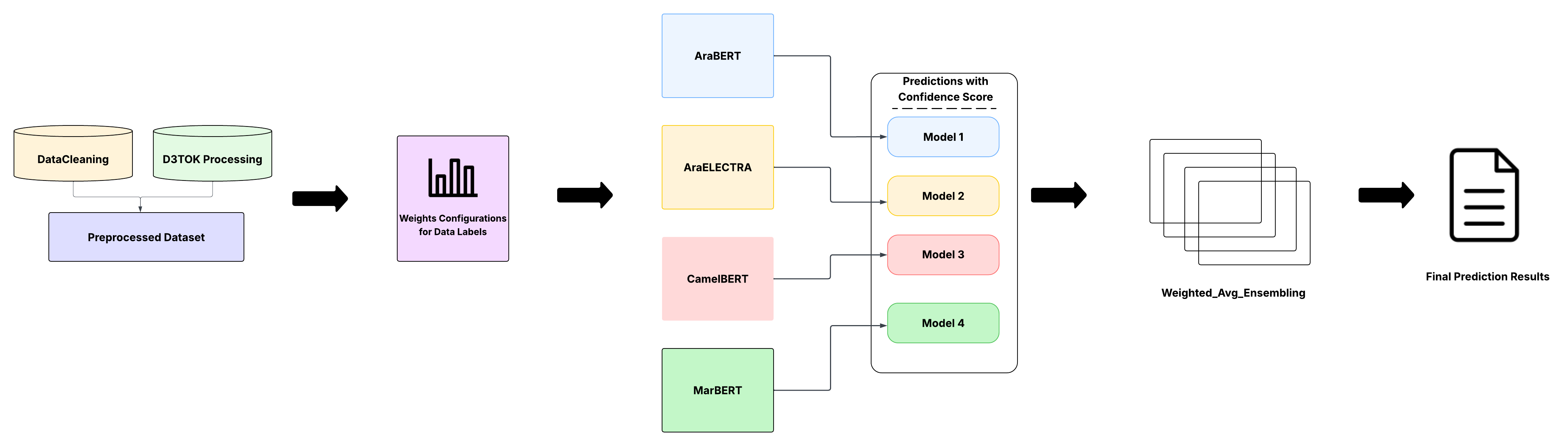}
  \caption{System Architecture Diagram}
  \label{fig:architecture}
\end{figure}

\subsection{Dataset Distribution} \label{app:readability_distribution}

\begin{figure}[H]
  \includegraphics[width=\linewidth]{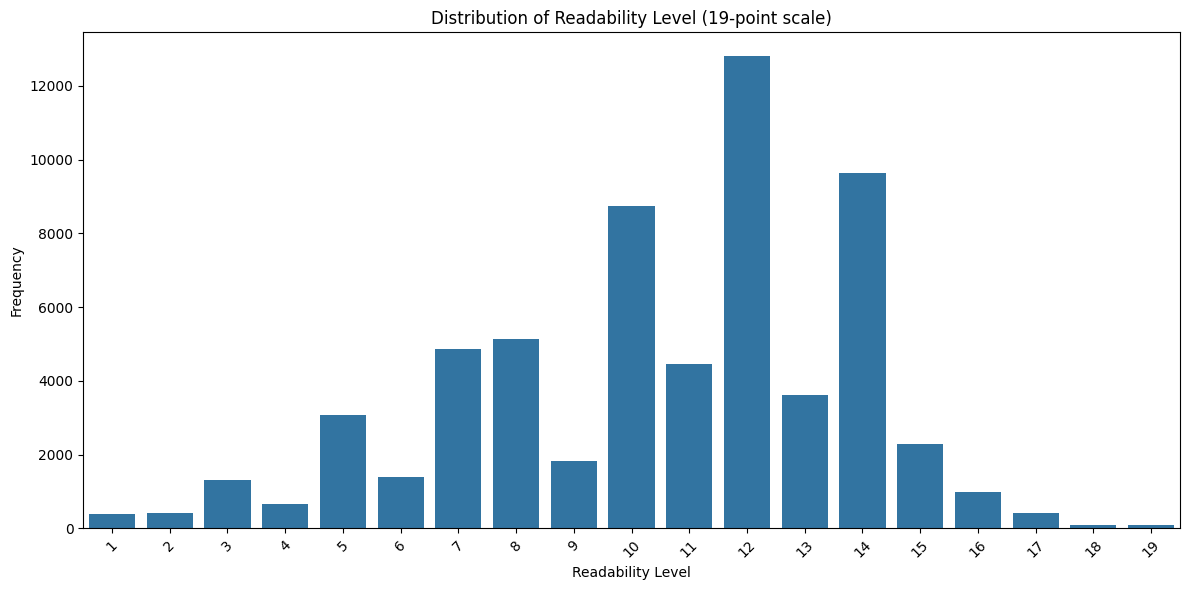}
  \caption{Distribution of Readability Levels across the dataset.}
  \label{fig:readability_distribution}
\end{figure}

\subsection{Training Configuration} \label{app:hyperparams}

\begin{table}[H]
\centering
\renewcommand{\arraystretch}{1.2}
\begin{tabular}{p{5cm}|p{5cm}}
\toprule
\textbf{Hyper-parameters} & \textbf{Values} \\
\midrule
Batch Size & 16 \\
Learning Rate & $2 X 10^-5$ \\
Epochs & 5 \\
Optimizer & AdamW \\
Callbacks & EarlyStopping \\
\bottomrule
\end{tabular}
\caption{Hyper-parameters used in model training.}
\label{tab:hyperparams}
\end{table}

\section{Data Augmentation Details}
\subsection{LLM Prompt Template} \label{app:llm_prompt}

\centering
\begin{tcolorbox}[enhanced, colback=gray!5, colframe=black!50, width=\textwidth, title=Few-Shot LLM Prompt]
\refstepcounter{tcbcounter}\label{box:llm_prompt}
\begin{Arabic}
المهمة: أعد صياغة الجملة العربية مع الحفاظ على:\\
- نفس عدد الكلمات\\
- نفس علامة الترقيم\\
- نفس مستوى القراءة: \{readability\_desc\}\\
- نفس المعنى العام\\
- استخدام مفردات وتراكيب بنفس مستوى الصعوبة\\

مثال 1:\\
الجملة الأصلية: "ماجد"\\
عدد الكلمات: 1\\
مستوى القراءة: مستوى أساسي جداً - كلمات بسيطة ومألوفة\\
علامة الترقيم: ""\\

الجملة المعاد صياغتها: "فهد"\\

مثال 2:\\
الجملة الأصلية: "السنة الثامنة"\\
عدد الكلمات: 2\\
مستوى القراءة: مستوى متوسط مبكر - كلمات متنوعة وتراكيب متوسطة\\
علامة الترقيم: ""\\

الجملة المعاد صياغتها: "العام الثامن"\\

مثال 3:\\
الجملة الأصلية: "الأربعاء 21 يناير 1987"\\
عدد الكلمات: 4\\
مستوى القراءة: مستوى فوق المتوسط - مفردات متقدمة وتراكيب معقدة\\
علامة الترقيم: ""\\

الجملة المعاد صياغتها: "يوم الأربعاء 21 كانون الثاني 1987"\\

=======================\\
الجملة المطلوب إعادة صياغتها:\\

الجملة الأصلية: "\{sentence\}"\\
عدد الكلمات: \{word\_count\}\\
مستوى القراءة: \{readability\_desc\}\\
علامة الترقيم: "\{punctuation\}"\\

المطلوب:\\
أعد صياغة الجملة مع الحفاظ على نفس عدد الكلمات (\{word\_count\})، ونفس مستوى القراءة، ونفس المعنى، ونفس علامة الترقيم "\{punctuation\}" في النهاية.\\

قدم فقط الجملة المعاد صياغتها بدون أي شرح أو تعليق إضافي.\\
\end{Arabic}
\end{tcolorbox}


\begin{thebibliography}{27}
\providecommand{\natexlab}{\#1}

\bibitem[{Abdul-Mageed et~al.(2021)Abdul-Mageed, Elmadany, and Nagoudi}]{abdul-mageed-etal-2021-arbert}
Muhammad Abdul-Mageed, AbdelRahim Elmadany, and El~Moatez Billah Nagoudi. 2021.
\newblock \href{https://aclanthology.org/2021.acl-long.551/}{ARBERT \& MARBERT: Deep bidirectional transformers for Arabic}.
\newblock In \emph{Proceedings of the 59th Annual Meeting of the Association for Computational Linguistics and the 11th International Joint Conference on Natural Language Processing (Volume 1: Long Papers)}, pages 7088--7105, Online. Association for Computational Linguistics.

\bibitem[{Aho and Ullman(1972)}]{Aho:72}
Alfred~V. Aho and Jeffrey~D. Ullman. 1972.
\newblock \emph{The Theory of Parsing, Translation and Compiling}, volume~1.
\newblock Prentice-Hall, Englewood Cliffs, NJ.

\bibitem[{Al Khalil et~al.(2020)Al Khalil, Habash, and Jiang}]{al-khalil-etal-2020-large}
Muhamed Al Khalil, Nizar Habash, and Zhengyang Jiang. 2020.
\newblock \href{https://aclanthology.org/2020.lrec-1.373/}{A large-scale leveled readability lexicon for Standard Arabic}.
\newblock In \emph{Proceedings of the Twelfth Language Resources and Evaluation Conference}, pages 3053--3062, Marseille, France. European Language Resources Association.

\bibitem[{Alhafni et~al.(2024)Alhafni, Hazim, Pineros Liberato, Al Khalil, and Habash}]{alhafni-etal-2024-samer}
Bashar Alhafni, Reem Hazim, Juan David Pineros Liberato, Muhamed Al Khalil, and Nizar Habash. 2025.
\newblock The SAMER Arabic text simplification corpus.
\newblock In \emph{Proceedings of the 2024 Joint International Conference on Computational Linguistics, Language Resources and Evaluation (LREC-COLING 2024)}, pages 16079--16093, Torino, Italia. ELRA and ICCL.

\bibitem[{{American Psychological Association}(1983)}]{APA:83}
{American Psychological Association}. 1983.
\newblock \emph{Publications Manual}.
\newblock American Psychological Association, Washington, DC.

\bibitem[{Ando and Zhang(2005)}]{Ando2005}
Rie Kubota Ando and Tong Zhang. 2005.
\newblock A framework for learning predictive structures from multiple tasks and unlabeled data.
\newblock \emph{Journal of Machine Learning Research}, 6:1817--1853.

\bibitem[{Andrew and Gao(2007)}]{andrew2007scalable}
Galen Andrew and Jianfeng Gao. 2007.
\newblock Scalable training of L1-regularized log-linear models.
\newblock In \emph{Proceedings of the 24th International Conference on Machine Learning}, pages 33--40.

\bibitem[{Antoun et~al.(2020)Antoun, Baly, and Hajj}]{antoun-etal-2020-arabert}
Wissam Antoun, Fady Baly, and Hazem Hajj. 2020.
\newblock \href{https://aclanthology.org/2020.osact-1.2/}{AraBERT: Transformer-based model for Arabic language understanding}.
\newblock In \emph{Proceedings of the 4th Workshop on Open-Source Arabic Corpora and Processing Tools, with a Shared Task on Offensive Language Detection}, pages 9--15, Marseille, France. European Language Resource Association.

\bibitem[{Antoun et~al.(2021)Antoun, Baly, and Hajj}]{antoun-etal-2021-araelectra}
Wissam Antoun, Fady Baly, and Hazem Hajj. 2021.
\newblock \href{https://aclanthology.org/2021.wanlp-1.20/}{AraELECTRA: Pre-training text discriminators for Arabic language understanding}.
\newblock In \emph{Proceedings of the Sixth Arabic Natural Language Processing Workshop}, pages 191--195, Kyiv, Ukraine (Virtual). Association for Computational Linguistics.

\bibitem[{Cao et~al.(2019)Cao, Mirjalili, and Raschka}]{cao2021coral}
Wenzhi Cao, Vahid Mirjalili, and Sebastian Raschka. 2019.
\newblock CORAL: Rank-consistent ordinal regression for neural networks.
\newblock \emph{arXiv preprint arXiv:1901.07884}.

\bibitem[{Chandra et~al.(1981)Chandra, Kozen, and Stockmeyer}]{Chandra:81}
Ashok~K. Chandra, Dexter~C. Kozen, and Larry~J. Stockmeyer. 1981.
\newblock Alternation.
\newblock \emph{Journal of the Association for Computing Machinery}, 28(1):114--133.

\bibitem[{El-Haj et~al.(2024)El-Haj, Almujaiwel, Premasiri, Ranasinghe, and Mitkov}]{el-haj-etal-2024-dares}
Mo El-Haj, Sultan Almujaiwel, Damith Premasiri, Tharindu Ranasinghe, and Ruslan Mitkov. 2024.
\newblock \href{https://aclanthology.org/2024.determit-1.10/}{DARES: Dataset for Arabic readability estimation of school materials}.
\newblock In \emph{Proceedings of the Workshop on DeTermIt! Evaluating Text Difficulty in a Multilingual Context @ LREC-COLING 2024}, pages 103--113, Torino, Italia. ELRA and ICCL.

\bibitem[{Elmadani et~al.(2025{\natexlab{a}})Elmadani, Alhafni, Taha, and Habash}]{elmadani-etal-2025-barec-shared-task}
Khalid~N. Elmadani, Bashar Alhafni, Hanada Taha, and Nizar Habash. 2025{\natexlab{a}}.
\newblock BAREC shared task 2025 on Arabic readability assessment.
\newblock In \emph{Proceedings of the Third Arabic Natural Language Processing Conference}, Suzhou, China. Association for Computational Linguistics.

\bibitem[{Elmadani et~al.(2025{\natexlab{b}})Elmadani, Habash, and Taha-Thomure}]{elmadani-etal-2025-large}
Khalid~N. Elmadani, Nizar Habash, and Hanada Taha-Thomure. 2025{\natexlab{b}}.
\newblock \href{https://aclanthology.org/2025.findings-acl.842/}{A large and balanced corpus for fine-grained Arabic readability assessment}.
\newblock In \emph{Findings of the Association for Computational Linguistics: ACL 2025}, pages 16376--16400, Vienna, Austria. Association for Computational Linguistics.

\bibitem[{Gusfield(1997)}]{Gusfield:97}
Dan Gusfield. 1997.
\newblock \emph{Algorithms on Strings, Trees and Sequences}.
\newblock Cambridge University Press, Cambridge, UK.

\bibitem[{Habash et~al.(2025)Habash, Taha-Thomure, Elmadani, Zeino, and Abushmaes}]{habash-etal-2025-guidelines}
Nizar Habash, Hanada Taha-Thomure, Khalid~N. Elmadani, Zeina Zeino, and Abdallah Abushmaes. 2025.
\newblock \href{https://aclanthology.org/2025.law-1.30/}{Guidelines for fine-grained sentence-level Arabic readability annotation}.
\newblock In \emph{Proceedings of the 19th Linguistic Annotation Workshop (LAW-XIX-2025)}, pages 359--376, Vienna, Austria. Association for Computational Linguistics.

\bibitem[{Inoue et~al.(2021)Inoue, Alhafni, Baimukan, Bouamor, and Habash}]{inoue-etal-2021-interplay}
Go Inoue, Bashar Alhafni, Nurpeiis Baimukan, Houda Bouamor, and Nizar Habash. 2021.
\newblock \href{https://aclanthology.org/2021.wanlp-1.10/}{The interplay of variant, size, and task type in Arabic pre-trained language models}.
\newblock In \emph{Proceedings of the Sixth Arabic Natural Language Processing Workshop}, pages 92--104, Kyiv, Ukraine (Virtual). Association for Computational Linguistics.

\bibitem[{Khered et~al.(2022)Khered, Abdelhalim, and Batista-Navarro}]{khered-etal-2022-building}
Abdullah Khered, Ingy Abdelhalim Abdelhalim, and Riza Batista-Navarro. 2022.
\newblock \href{https://aclanthology.org/2022.wanlp-1.53/}{Building an ensemble of transformer models for Arabic dialect classification and sentiment analysis}.
\newblock In \emph{Proceedings of the Seventh Arabic Natural Language Processing Workshop (WANLP)}, pages 479--484, Abu Dhabi, United Arab Emirates (Hybrid). Association for Computational Linguistics.

\bibitem[{Liberato et~al.(2024)Liberato, Alhafni, Khalil, and Habash}]{liberato-etal-2024-strategies}
Juan Liberato, Bashar Alhafni, Muhamed Khalil, and Nizar Habash. 2024.
\newblock \href{https://aclanthology.org/2024.arabicnlp-1.5/}{Strategies for Arabic readability modeling}.
\newblock In \emph{Proceedings of the Second Arabic Natural Language Processing Conference}, pages 55--66, Bangkok, Thailand. Association for Computational Linguistics.

\bibitem[{Obeid et~al.(2020)Obeid, Zalmout, Khalifa, Taji, Oudah, Alhafni, Inoue, Eryani, Erdmann, and Habash}]{obeid-etal-2020-camel}
Ossama Obeid, Nasser Zalmout, Salam Khalifa, Dima Taji, Mai Oudah, Bashar Alhafni, Go Inoue, Fadhl Eryani, Alexander Erdmann, and Nizar Habash. 2020.
\newblock \href{https://aclanthology.org/2020.lrec-1.868/}{CAMeL tools: An open source Python toolkit for Arabic natural language processing}.
\newblock In \emph{Proceedings of the Twelfth Language Resources and Evaluation Conference}, pages 7022--7032, Marseille, France. European Language Resources Association.

\bibitem[{Rasooli and Tetreault(2015)}]{rasooli-tetrault-2015}
Mohammad Sadegh Rasooli and Joel~R. Tetreault. 2015.
\newblock \href{http://arxiv.org/abs/1503.06733}{Yara parser: A fast and accurate dependency parser}.
\newblock \emph{Computing Research Repository}, arXiv:1503.06733, version~2.

\end{thebibliography}
\end{document}